\documentclass[runningheads]{llncs}
\usepackage{graphicx}
\usepackage{amsmath,amssymb} % define this before the line numbering.
\usepackage{color}
\usepackage{url}
\begin{document}
	\title{A Top-down Approach to Articulated Human Pose Estimation and Tracking}
	% Replace with your title
	
	\titlerunning{Articulated Pose Estimation and Tracking}
	% Replace with a meaningful short version of your title
	
	\author{Guanghan Ning\orcidID{0000-0002-4356-7862} \and
		Ping Liu\orcidID{0000-0002-3170-3783} \and
		Xiaochuan Fan\orcidID{0000-0002-5346-2925} \and
		Chi Zhang\orcidID{0000-0001-8409-1189}}
	%Please write out author names in full in the paper, i.e. full given and family names. 
	%If any authors have names that can be parsed into FirstName LastName in multiple ways, please include the correct parsing, in a comment to the volume editors:
	%\index{Lastnames, Firstnames}
	%(Do not uncomment it, because you may introduce extra index items if you do that, we will use scripts for introducing index entries...)
	\authorrunning{G. Ning, P. Liu, X.Fan and C.Zhang}
	% Replace with shorter version of the author list. If there are more authors than fits a line, please use A. Author et al.
	\institute{JD.com Silicon Valley Research Center \\
	\email{\{guanghan.ning, ping.liu, xiaochuan.fan, chi.zhang\}@jd.com}}

\maketitle              % typeset the header of the contribution
\begin{abstract}
	Both the tasks of multi-person human pose estimation and pose tracking in videos are quite challenging. Existing methods can be categorized into two groups: top-down and bottom-up approaches. In this paper, following the top-down approach, we aim to build a strong baseline system with three modules: human candidate detector, single-person pose estimator and human pose tracker. Firstly, we choose a generic object detector among state-of-the-art methods to detect human candidates. Then, cascaded pyramid network is used to estimate the corresponding human pose.  Finally, we use a flow-based pose tracker to render keypoint-association across frames, i.e., assigning each human candidate a unique and temporally-consistent id, for the multi-target pose tracking purpose. We conduct extensive ablative experiments to validate various choices of models and  configurations. We take part in two ECCV'18 PoseTrack challenges\footnote{ \url{https://posetrack.net/workshops/eccv2018/posetrack_eccv_2018_results.html}}:  pose estimation and pose tracking. 

\end{abstract}

\section{Introduction}
\label{sec:intro}
Compared to single person human pose estimation, where human candidates are cropped and centered in the image patch, the task of multi-person human pose estimation is more realistic and challenging. Existing methods can be classified into top-down and bottom-up approaches. The top-down approach \cite{fang2017rmpe,papandreou2017towards} relies on a detection module to obtain human candidates and then apply a single-person human pose estimator to locate human keypoints. 
The bottom-up approach  \cite{cao2016realtime,he2017mask,xia2017joint,newell2016associative}, on the other hand, detects human keypoints from all potential human candidates and then assembles these keypoints into human limbs for each individual based on various data association techniques. 
The advantage of bottom-up approaches is their excellent trade-off between estimation accuracy and computational cost because their computational cost is invariant to the number of human candidates in the image. 
In contrast, the main advantage of top-down approaches is their capability in disassembling the task into multiple comparatively easier tasks, i.e., object detection and single-person pose estimation. The object detector is expert in detecting hard (usually small) candidates, so that the pose estimator will perform better with a focused regression space.

Pose tracking is the task of estimating human keypoints and assigning unique ids for each keypoint at instance-level across frames in videos. In videos with multiple people, accurate trajectory estimation of human key points is useful in human action recognition and human interaction understanding. 
PoseTrack\cite{iqbal2016posetrack} and ArtTrack\cite{insafutdinov2017arttrack} primarily introduce multi-person pose tracking challenge and propose a graph partitioning formulation, which transforms the pose tracking problem into a minimum cost multi-cut problem. 
However, hand-crafted graphical models are not scalable for long and unseen clips. 
Another line of research explores top-down approach \cite{girdhar2018detect,xiu2018pose,xiao2018simple} by operating multi-person human pose estimation on each frame and linking them based on appearance similarities and temporal adjacencies. A naive solution is to apply multi-target object tracking on human detection candidates across frames and then estimate human poses for each human tubelet. While this is a feasible method, it neglects unique attributes of keypoints. Compared to the tracked bounding boxes, keypoints can potentially be helpful cues for both the bounding boxes and the keypoints tracking. The tracker of 3D Mask R-CNN \cite{girdhar2018detect} simplifies the pose tracking problem as a maximum weight bipartite matching problem and solve it with Greedy or Hungarian algorithm. PoseFlow \cite{xiu2018pose} further takes motion and pose information into account to address the issue of occasional truncated human candidates.

\section{Our Approach}
\label{sec:methodology}
We follow the top-down approach for pose tracking, i.e., perform human candidate detection, single-person pose estimation, and pose tracking step by step. The details for each module are described below, respectively. 

\subsection{Detection Module}
We adopt state-of-the-art object detectors trained with ImageNet and COCO datasets. Specifically, we use pre-trained models from deformable ConvNets \cite{dai17dcn}. 
In order to increase the recall rate of human candidates, we conduct experiments on validation sets of both PoseTrack 2017 \cite{andriluka2018posetrack} and PoseTrack 2018 to choose the best object detector.
Firstly, we infer ground truth bounding boxes of human candidates from the annotated keypoints, because in PoseTrack 2017 dataset, the bounding box position is not provided in the annotations. Specifically, we locate a bounding box from the minimum and maximum coordinates of the 15 keypoints, and then enlarge this box by 20\% both horizontally and vertically. Even though ground truth bounding boxes are given in PoseTrack 2018 dataset, we infer a more consistent version based on ground truth locations of keypoints. 
Those inferred ground truth bounding boxes are utilized to train the pose estimator.

For the object detectors, we compare the deformable convolution versions of the R-FCN network \cite{dai2016r} and of the FPN network \cite{lin2017feature}, both with ResNet101 backbone\cite{he2016deep}.
The FPN feature extractor is attached to the Fast R-CNN\cite{girshick2015fast} head for detection.
We compare the detection results with the ground truth based on the precision and recall rate on PoseTrack 2017 validation set.  
In order to eliminate redundant candidates, we drop candidate(s) with lower {likelihood}. 
As shown in Table \ref{table-detectors}, for various drop thresholds of bounding boxes, the precision and recall of the detectors are given. For PoseTrack 2018 validation set, the FPN network performs better as well. Therefore, we choose the FPN network as our human candidate detector. 

\begin{table}
	\caption{Precision-Recall on PoseTrack 2017 validation set. A bounding box is correct if its IoU with GT is above certain threshold, which is set to 0.4 for all experiments.}\label{table-detectors}
	\centering 
	\begin{tabular}{l|l|l|l|l|l|l|l|l|l}
		\hline
		Drop thresholds of bbox &  0.1 & 0.2 & 0.3 & \textbf{0.4} & 0.5 & 0.6 & 0.7 & 0.8 & 0.9 \\
		\hline
		Deformable FPN (ResNet101): prec & 17.9 & 27.5 &32.2 &\textbf{34.2}  &35.7 &37.2 &38.6 &40.0 &42.1 \\
		Deformable R-FCN (ResNet101): prec & 15.4 & 21.1 &25.9 &30.3  &34.5 &37.9 &39.9 &41.6 &43.2 \\
		\hline
		Deform FPN (ResNet101): recall & 87.7 &86.0 &84.5  &\textbf{83.0} &80.8 &79.2 &77.0 &73.8 &69.0 \\
		Deform R-FCN (ResNet101): recall & 87.7 &86.5 &85.0  &82.6 &80.1 &77.3 &74.4 &70.4 &61.0 \\
		\hline
	\end{tabular}
\end{table}

The upper bound for detection is the ground truth bounding box location. In order to measure the gap between ideal detection results and our detection results, we feed the ground truth bounding boxes to the subsequent pose estimation module and tracking module, and compare its performance with that of our detector on the validation set.
As shown in Table \ref{table-gap-task1}, the pose estimation will  perform around 7\% better with ground truth detections.  
As shown in Table \ref{table-gap-task3}, the pose tracking will perform around 6\% better with ground truth detections.  

\begin{table}
	\caption{Comparison of single-frame pose estimation results using various detectors on PoseTrack 2017 validation set. }\label{table-gap-task1}
	\centering 
	\begin{tabular}{l|l|l|l|l|l|l|l|l}
		\hline
		Average Precision (AP)       & Head & Shou & Elb  & Wri  & Hip  & Knee & Ankl & \textbf{Total} \\
		\hline
		Ground Truth Detections & 88.9 & 88.4 & 82.7 & 74.7 & 78.9 & 79.4 & 75.4 & \textbf{81.7} \\
		Deform FPN (ResNet101)  & 80.7 & 81.2 & 77.4 & 70.2 & 72.6 & 72.2 & 64.7 & \textbf{74.6} \\
		Deform R-FCN (ResNet101) & 79.6 & 80.3 & 75.9 & 69.0 & 72.0 & 71.6 & 64.3 & \textbf{73.7} \\
		\hline
	\end{tabular}
\end{table}

\begin{table}
	\caption{Comparison of multi-frame pose tracking results using various detectors on PoseTrack 2017 validation set. }\label{table-gap-task3}
	\centering 
	\begin{tabular}{l|l|l|l|l|l|l|l|l}
		\hline
		- &MOTA & MOTA & MOTA & MOTA & MOTA & MOTA & MOTA & \textbf{MOTA}  \\
		- &Head & Shou & Elb  & Wri  & Hip  & Knee & Ankl & \textbf{Total}\\
		\hline
		GT Detections & 78.8 & 78.2 & 65.6 & 56.3 & 64.4 & 63.8 & 56.2 & \textbf{67.0} \\
		D-FPN-101  & 68.9 & 70.9 & 62.7 & 54.6 & 59.5 & 59.8 & 48.7 & \textbf{61.3} \\
		D-RFCN-101 & 66.5 & 68.1 & 60.1 & 52.2 & 57.4 & 57.9 & 47.4 & \textbf{59.0} \\
		\hline
	\end{tabular}
\end{table}

With ResNet151 as backbone, and training detectors solely on the human class, e.g., training on the CrowdHuman\cite{shao2018crowdhuman} dataset, we believe the detection module may render better results. For the challenge, we just adopt the deformable FPN with ResNet101 and use their pre-trained model for simplicity.

\subsection{Pose Estimation Module}
For the single-person human pose estimator, we adopt Cascaded Pyramid Networks (CPN) \cite{chen2017cascaded} with slight modifications. 
We first train the CPN network with the merged dataset of PoseTrack 2018 and COCO for 260 epochs. Then we finetune the network solely on PoseTrack 2018 training set for 40 epochs in order to mitigate the regression on head. For COCO dataset, bottom-head and top-head positions are not given. We infer these keypoints through rough interpolation on the annotated keypoints. We find that by finetuning on the PoseTrack dataset, the prediction on head keypoints will be refined.  During finetuning, we use the technique of online hard keypoint mining, only focusing on losses from the 7 hardest keypoints out of the total 15 keypoints. 

In our implementation, we perform non-maximum suppression (NMS) in the detection phase on the bounding boxes and perform pose estimation on all candidates from the detection module. For each candidate, we post-process on the predicted heatmaps with cross-heatmap pose NMS \cite{ning2018knowledge} to render more accurate keypoint locations. We did not perform flip testing, although the performance might be slightly better. 
During testing, we use a manifold of two models from epoch 291 and 293. We notice a slight performance boost with model ensemble. For epoch 291, the prediction of shoulders and hips renders better results than epoch 293 on validation sets of both PoseTrack 2017 and PoseTrack 2018. However, epoch 293 performs better on end limbs such as ankles and wrists. We test with two manifold modes: (1) Average and (2) Expert. As shown in Table \ref{table-manifold}, the expert mode takes shoulder/hip predictions from the previous model and end-limb predictions from the latter, which performs better consistently on both PoseTrack 2017 and PoseTrack 2018 validation sets. Both modes perform better than plain testing on the pose estimation task.

\begin{table}
	\caption{Comparison of single-frame pose estimation results with different ensemble modes on PoseTrack 2017 validation set. }\label{table-manifold}
	\centering 
	\begin{tabular}{l|l|l|l|l|l|l|l|l}
		\hline
		Average Precision (AP)       & Head & Shou & Elb  & Wri  & Hip  & Knee & Ankl & \textbf{Total} \\
		\hline
		Epoch 291           & 80.7 & \textbf{81.2} & 77.4 & 70.2 & \textbf{72.6} & 72.2 & 64.7 & 74.6 \\
		Epoch 293          & 80.5 & 80.8 & 77.9 & \textbf{71.3} & 70.1 & 72.9 & \textbf{65.7} & 74.6 \\
		Average     & 81.3 & 81.2 & 77.6 & 70.7 & 72.1 & 72.5 & 65.1 & 74.8 \\
		Expert       & 80.6 & 81.2 & 77.9 & 71.3 & 72.6 & 72.9 & 65.7 & 75.0 \\
		\hline
	\end{tabular}
\end{table}

\subsection{Pose Tracking Module}

\begin{figure}
	\centering
	\includegraphics[width=0.7\linewidth]{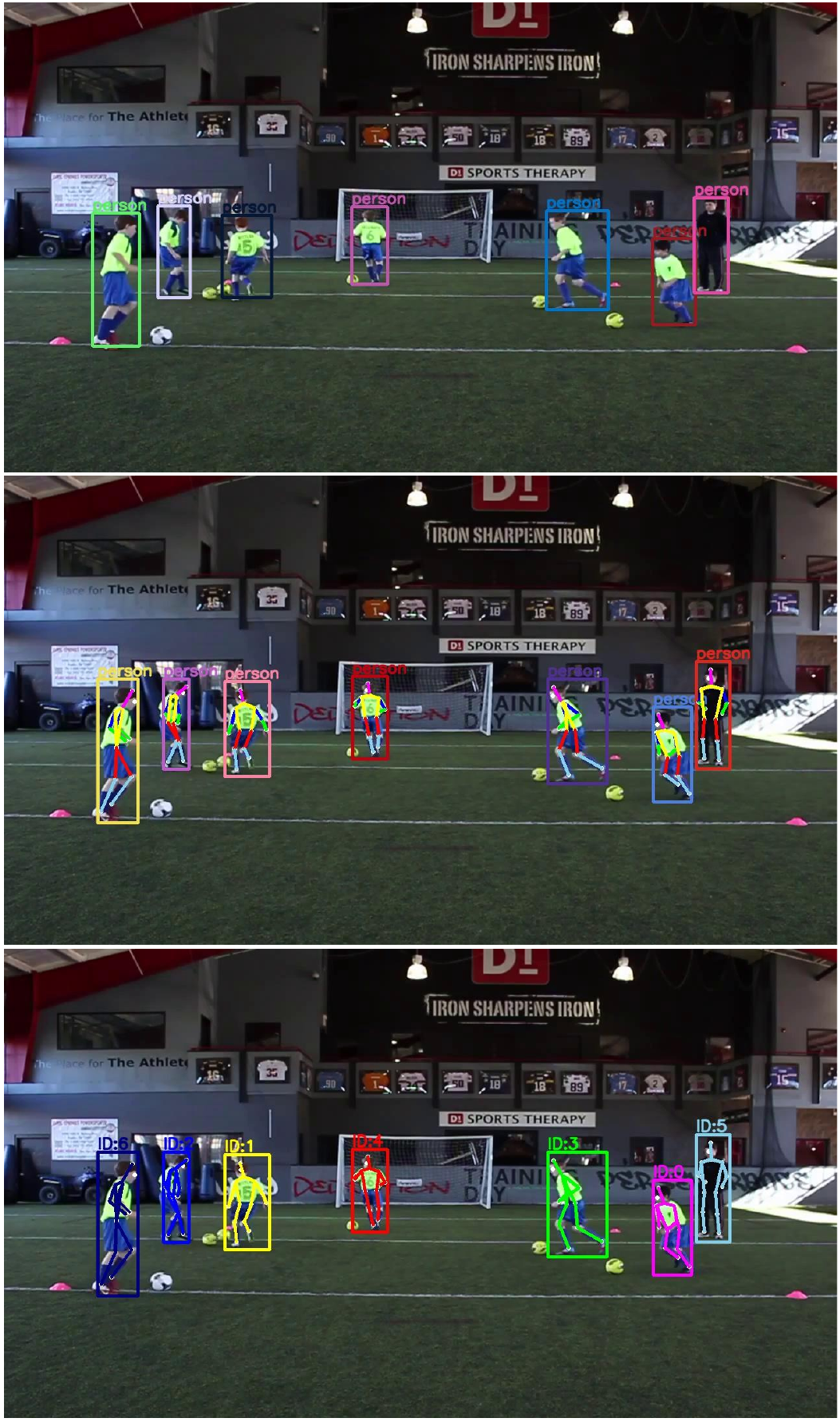}
	\caption{Our modular system for pose tracking. From top to bottom: we perform human candidate detection, pose estimation, and pose tracking sequentially.} 
	\label{fig:overview}
\end{figure}

We adopt a flow-based pose tracker \cite{xiu2018pose}, where pose flows are built by associating poses that indicate the same person across frames.
We start the tracking process from the first frame where human candidates are detected. 
To prevent assignments of IDs for persons which have already left the visible image area, IDs are only kept for a limited amount of frames, afterwards they are discarded.
For the pose tracking task, the performance is evaluated via MOTA, which is very strict. It penalizes mis-matches, false positives and misses. In order to get higher MOTA results, we need to drop keypoints with lower confidence scores, sacrificing the recall rate of correct keypoints.
We find the MOTA evaluation criterion quite sensitive to the drop rate of keypoints, as shown in Table \ref{table-MOTA-drop}.
\begin{table}
	\caption{ Sensitivity analysis on how the drop thresholds of keypoints affect the performance in AP and MOTA. Performed on PoseTrack 2018 validation set. }\label{table-MOTA-drop}
	\centering 
	\begin{tabular}{l|l|l|l|l|l}
		\hline
		Threshold       & 0.5 & 0.6 & 0.7 & 0.8  & 0.85  \\
		\hline
		Pose Estimation (AP)     & \textbf{76.3} & 75.5 & 73.4 & 69.7 & 67.1 \\
		Pose Tracking (MOTA)     & 40.4 & 53.4 & 60.6  & \textbf{62.4} & 61.6  \\
		\hline
	\end{tabular}
\end{table}

Considering the distinct difficulties of keypoints, e.g., shoulders are easier than ankles to localize, the confidence distribution for each joint is supposedly not uniform. Dropping keypoints solely based on the keypoint confidence estimated by the pose estimator may not be an ideal strategy for pose tracking. We collect statistics on the drop rate of keypoints from different joints, as shown in Table \ref{table-statistics}.
We can see that from left to right, the keypoints become more and more difficult to estimate, as reflected by their respective preservation rate. 
The least and most difficult joints are the shoulders and ankles, respectively. In other words, the pose estimator is most confident on the shoulders but least confident on ankles. 
An adaptive keypoint pruner may help increase the MOTA performance while maintaining high recall rates.

\begin{table}
	\caption{ Statistics analysis on the drop rates of keypoints with different drop thresholds. Performed on PoseTrack 2018 validation set. The numbers indicate the percentage of keypoints maintained after pruning. }\label{table-statistics}
	\centering 
	\begin{tabular}{l|l|l|l|l|l|l|l|l}
		\hline
		Threshold       & Shou & Head & Elb & Hip & Knee  & Wri & Ankl  & Total \\
		\hline
		0.70     & 82.1 & 75.3 & 68.3  & 66.0 & 60.2 &60.2 &54.6 &68.6 \\
		0.75    & 78.4 & 71.1 & 63.9  & 61.5  & 56.2 &54.9 &49.9 &64.3 \\
		0.85    & 70.2 & 62.3 & 54.3 & 53.0 & 48.8 &46.2 & 42.3 &56.0 \\
		\hline
	\end{tabular}
\end{table}

\section{Challenge Results}
\label{sec:exp}
Our final performance on the partial test set of PoseTrack 2018 is given in Table \ref{table-results-task2} and Table \ref{table-results-task3}.

\begin{table}
	\caption{Our single-frame pose estimation results on PoseTrack 2018 partial test set}\label{table-results-task2}
	\centering 
	\begin{tabular}{l|l|l|l|l|l|l|l|l}
		\hline
		Average Precision (AP)       & Head & Shou & Elb  & Wri  & Hip  & Knee & Ankl & \textbf{Total} \\
		\hline
		Ours     & 74.2 & 74.3 & 71.5 & 66.8 & 66.7 & 67.2 & 62.4 & \textbf{69.4} \\
		\hline
	\end{tabular}
	
\end{table}
\begin{table}
	\caption{Our multi-frame pose tracking results on PoseTrack 2018 partial test set}\label{table-results-task3}
	\centering 
	\begin{tabular}{l|l|l|l|l|l|l|l|l|l|l|l}
		\hline
		- & MOTA & MOTA & MOTA & MOTA & MOTA & MOTA & MOTA & \textbf{MOTA} & MOTP & Prec & Rec  \\
		- & Head & Shou & Elb  & Wri  & Hip  & Knee & Ankl & \textbf{Total}& Total& Total& Total\\
		\hline
		Ours & 60.2 & 62.1 & 53.9 & 50.1 & 52.2 & 52.6 & 47.4 & \textbf{54.5} & 85.9 & 83.9 & 68.9 \\ 
		\hline
	\end{tabular}
\end{table}

%=================================================================
\section{Conclusion}
\label{sec:conclusion}
In this paper, we aim to build a modular system to reach the state-of-the-art of human pose estimation and tracking. This system consists of three modules, which conduct human candidate detection, pose estimation and pose tracking respectively. 
We have analyzed each module in the system with ablation studies on various models and configurations while discussing their pros and cons. 
We present the performance of our system in the pose estimation challenge and pose tracking challenge of PoseTrack 2018.


\newpage
\bibliographystyle{splncs04}
\bibliography{ning}

\end{document}